\documentclass{IOS-Book-Article}

\usepackage{mathptmx}
\usepackage{graphicx}
\usepackage{booktabs}
\usepackage{xcolor}
\usepackage{amsmath}

\makeatletter
\providecommand{\@LN@col}[1]{}
\providecommand{\@LN}[2]{}
\makeatother

\newif\ifreviewcopy
\reviewcopyfalse

\begin{document}
\raggedbottom
\begin{frontmatter}

\title{Legal Guidance in the Wild: How Users Seek Legal Help in Real-World LLM Conversations}

\author[A]{\fnms{Amirreza} \snm{Mohseni}},
\author[A]{\fnms{Gerasimos} \snm{Spanakis}}
\runningauthor{A. Mohseni and G. Spanakis}
\address[A]{Maastricht University, Maastricht, The Netherlands}

\begin{abstract}
People ask conversational AI models about their legal rights, obligations, procedures, and personal situations that may have legal consequences. To study this behavior, we first need to distinguish requests for legal guidance from other legally themed discussions and unrelated conversations. We contribute a multi-axis taxonomy of legal queries capturing whether legal guidance is sought, how a request is formulated, and which legal domains it concerns; apply it to 2,122 naturally occurring WildChat-1M conversations; and create a human-reviewed gold evaluation set of 200 conversations. All annotations, model checkpoints, and evaluation code are publicly available. In this enriched sample, 71.1\% of conversations containing
legal content seek personalized or practical guidance. Rights-and-options advice
is the most common primary request type among legal-content
conversations, and 25.3\% of guidance-positive conversations
span multiple legal topics. The gold evaluation set supports a two-stage comparison of ModernBERT-base and ModernBERT-large against a TF--IDF cascade. Our framework supports large-scale studies of legal guidance in conversational AI, and the public release of the annotated data facilitates reproducible research and future model development.
\end{abstract}

\begin{keyword}
legal natural language processing\sep conversational AI\sep legal guidance seeking
\end{keyword}

\end{frontmatter}

\begingroup
\renewcommand{\thefootnote}{}
\footnotetext{Accepted for publication at JURIX 2026, Toulouse, France.}
\endgroup

\ifreviewcopy
  \pagestyle{plain}
  \thispagestyle{plain}
\fi

\section{Introduction}

People use conversational AI to seek both legal information and advice about what action to take \cite{kuk2025llmslegalaid}. In our dataset, these questions span topics such as ``Immigration \& Citizenship,'' ``Employment \& Labor,'' ``Housing \& Property,'' ``Business \& Commercial,'' and ``Consumer \& Personal Finance.'' Prominent legal NLP benchmarks evaluate legal-document understanding, retrieval, answer generation, or expert-defined reasoning tasks \cite{chalkidis2022lexglue,guha2023legalbench,louis2022statutory,louis2024interpretable}. They reveal less about the more basic question faced by a general-purpose assistant: when is a person actually seeking legal guidance, and how are such requests formulated in the wild?

Legally themed text is not synonymous with guidance seeking. A user may ask for a statute to be translated, discuss a fictional dispute, request an academic explanation, or describe a personal situation while seeking concrete rights, options, or next steps. We therefore study a hierarchical classification problem over real LLM conversations: the first stage predicts whether the user seeks personalized or practical legal guidance, and the second assigns each positive conversation to one of 14 primary legal topics.

The distinction matters because public-facing LLMs can confidently produce unreliable legal information \cite{dahl2024legalfictions}: a compact guidance detector could support large-scale extraction for empirical research or, after calibration and user testing, trigger a notice that generated output is not a substitute for professional advice. This framing parallels recent work on how users seek health information from conversational systems \cite{paruchuri2025whatsupdoc}.

We make four contributions. First, we develop a taxonomy for identifying and characterizing requests for legal guidance in multi-turn conversations. Second, we apply it to 2,122 naturally occurring WildChat-1M conversations spanning both guidance-seeking and non-guidance cases and analyze how requests are formulated, when they first arise, and which legal domains they concern. Third, we release a human-reviewed gold evaluation set of 200 conversations and supporting artifacts.\footnote{\url{https://github.com/Amir-Mohseni/Legal-Guidance-in-the-Wild}} Fourth, we compare a lexical cascade with ModernBERT-base and ModernBERT-large \cite{warner2024modernbert}, evaluating them on the gold evaluation set only after finalizing all modeling decisions using separate silver-labeled development data.

\section{Related Work}

\noindent\textbf{Legal information needs.} Prior work has examined both observed requests to legal-help AI and users' expectations of such systems. Kuk and Hara{\v{s}}ta study queries submitted to a GPT-4 legal assistant and distinguish requests for legal information from requests for advice about what action to take \cite{kuk2025llmslegalaid}. Hagan uses interviews and design sessions to examine when people would use legal-help AI and which qualities and safeguards they expect from it \cite{hagan2024standards}. Steenhuis and Westermann use LLMs to support civil legal-aid intake by checking eligibility and generating follow-up questions \cite{steenhuis2024getting}. We study an upstream and broader task in naturally occurring conversations: identifying whether users seek legal guidance and how they formulate those requests.

\noindent\textbf{Behavior in real-world conversations.} WildChat makes it possible to study interactions with conversational models at scale \cite{zhao2024wildchat}. A close methodological analogue is \emph{What's Up, Doc?}, which filters large conversational datasets and applies a clinician-driven taxonomy to how and why users seek health information across specialties \cite{paruchuri2025whatsupdoc}. We adopt the same user-centered perspective for law, but separate legal content from guidance seeking and add hierarchical topic classification, supporting both descriptive analysis and smaller classifiers that apply the taxonomy without repeatedly invoking a frontier model.

\noindent\textbf{Legal NLP.} Prominent benchmarks evaluate legal-document understanding or expert-defined reasoning tasks \cite{chalkidis2022lexglue,guha2023legalbench}. A complementary line of work starts from citizen or lay legal questions. BSARD maps French questions to relevant Belgian statutory provisions, and \emph{Finding the Law} studies improved statutory retrieval in the same citizen-question setting \cite{louis2022statutory,louis2023finding}. LLeQA pairs expert-annotated questions with long-form answers grounded in legislation \cite{louis2024interpretable}, while GerLayQA pairs German civil-law questions from laypeople with lawyer-authored answers and statutory references \cite{buttner2024answering}. These resources provide useful evidence about question language, legal-topic distributions, retrieval, and answer generation in fixed-question settings. We instead examine the distributions of guidance intent, request types, and legal topics in naturally occurring, multi-turn conversations. Topic taxonomies provide useful structure for legal datasets \cite{sargeant2026taxonomy}, but assigning an interaction directly to a legal field misses whether the user wants explanation, drafting, procedural steps, or advice about rights and options. Strong benchmark performance also does not make a public model a reliable legal adviser: Dahl et al.\ document frequent legal hallucinations and confident acceptance of false premises \cite{dahl2024legalfictions}. Our classifier does not assess answer correctness; it identifies conversations in which stronger safeguards, specialized routing, or human escalation may be warranted.

\section{Data and Task}
\label{sec:data}

\subsection{Dataset construction, taxonomy, and silver annotation}

The conversations originate from English-language WildChat-1M
interaction logs \cite{zhao2024wildchat}. We used OpenAI's
GPT-5.6 Terra, accessed through the OpenAI API at high reasoning
effort, for an initial screening pass to identify clearly
unrelated material and conversations potentially containing
legal content \cite{openai2026gpt56}. The retained sample contains 2,122 conversations, combining
conversations initially identified as containing legal content
with an equal-sized random sample from those classified by
Terra as non-legal. These initial labels
determined sample composition and are distinct from the
annotations used in our analysis. This design allows building classification models; however, the label distribution cannot
estimate prevalence in WildChat.

The taxonomy captures three aspects of a conversation.
Its \emph{legal relevance} axis distinguishes legal content
from personalized or practical guidance seeking.
Its \emph{interaction} axis records user context, drafting
intent, guidance onset, and request types: rights-and-options
advice, information or explanation, procedural assistance,
document drafting, document review, and other requests.
Its \emph{domain} axis assigns legal topics.

We adapted the topic and subtopic structure of existing literature 
\cite{louis2022statutory,louis2023finding}. The domain axis
concerns 13 substantive legal topics and a residual
``\textsc{Other}'' category. Each conversation receives one
primary topic representing its predominant legal issue and
optional secondary topics for additional issues. The domain
axis is flat; the classification pipeline first detects
guidance seeking and then predicts a primary topic.
The other taxonomy fields support descriptive analysis.

Following screening and sample selection, we annotated all
2,122 retained conversations with OpenAI's GPT-5.6 Sol,
accessed through the OpenAI API at medium reasoning effort
under the taxonomy \cite{openai2026gpt56}. Sol annotated every
retained conversation, including the non-guidance cases;
Terra's screening labels were not used as the final silver
labels. These Sol annotations form the silver-labeled dataset:
1,367 conversations contain legal content and 972 seek legal
guidance (45.8\%). All classifiers and descriptive analyses
use these annotations, with human-reviewed labels replacing
them for gold evaluation. Only user messages are presented,
in chronological order; assistant messages are excluded
because we study how users phrase legal requests.
The dataset, annotation prompt, JSON schema, and run
configurations are available through the project resources.
\footnote{\url{https://huggingface.co/collections/AmirMohseni/legal-guidance-in-the-wild}}

We retain a fixed 200-conversation subset for human review.
It was sampled using the initial guidance labels to include
100 guidance-positive and 100 guidance-negative conversations;
topic was not used for selection. After Sol annotation and
human review, this \emph{gold evaluation set} contains
84 guidance-positive and 116 guidance-negative conversations.
The other 1,922 conversations form the \emph{silver development
set}, split into 1,632 training and 290 validation conversations.
The split was stratified first by guidance and then, among
guidance-positive conversations where counts allowed, by primary
topic (Table~\ref{tab:dataset-summary}). The term
\emph{silver-labeled dataset} refers to all 2,122 Sol annotations,
including those of the evaluation conversations before human
review. Table~\ref{tab:topic-distribution} reports the complete
primary-topic distribution.

\begin{table}[t]
\centering
\footnotesize
\setlength{\tabcolsep}{5pt}
\begin{tabular}{lrrrr}
\toprule
Split & $n$ & Guidance & Legal content & Human guidance \\
\midrule
Silver-labeled dataset & 2,122 & 972 (45.8\%) & 1,367 (64.4\%) & -- \\
Silver development set & 1,922 & 884 (46.0\%) & 1,232 (64.1\%) & -- \\
\quad Train & 1,632 & 750 (46.0\%) & 1,048 (64.2\%) & -- \\
\quad Validation & 290 & 134 (46.2\%) & 184 (63.4\%) & -- \\
Gold evaluation set & 200 & 88 (44.0\%) & 135 (67.5\%) & 84 (42.0\%) \\
\bottomrule
\end{tabular}
\caption{Dataset and label summary. The Guidance and Legal content columns report silver labels; the last column reports the final human-reviewed guidance labels, which exist only for the gold evaluation set and replace the corresponding silver guidance labels during evaluation. The dataset was enriched during screening, so these percentages are not WildChat prevalence estimates.}
\label{tab:dataset-summary}
\end{table}

\subsection{Human review and gold evaluation set}

Two reviewers with backgrounds in computer science independently examined all 200 conversations in the gold evaluation set. Each reviewer worked on a separate worksheet and recorded whether the silver guidance and primary-topic labels were correct, adding a corrected label and short note where needed. Reviewing the silver labels rather than labeling each conversation from scratch allowed both reviewers to assess the full set, but may have introduced anchoring; we return to this in the Discussion.

We measured agreement on the 200-conversation gold evaluation
set using the available paired judgments. Guidance agreement
was 96.5\%, with Cohen's $\kappa = .927$. Topic agreement,
measured over binary judgments of whether the silver topic
was correct, was 94.9\%, with $\kappa = .561$. The lower
topic $\kappa$ reflects judgment imbalance: reviewers marked
the silver topic as correct in 94\% of judgments, and $\kappa$
is sensitive to such imbalances \cite{zec2017kappa}. Overall, the reviewers reached full agreement on 186 conversations (93.0\%). They discussed the remaining 14 cases (7.0\%) and reached consensus; these consensus decisions became the final labels. We use only the silver development set for training and model selection and reserve the gold evaluation set for final evaluation.

\subsection{Task and metrics}

We formulate the task as a two-stage classification pipeline. In Stage 1, the model identifies whether a conversation contains a request for legal guidance. In Stage 2, conversations identified as positive in Stage 1 are classified into one of 14 legal topics. Each conversation receives one of 15 final labels (either one of the 14 topics or ``\textsc{No Guidance}''), so errors at either stage affect the final label. We report accuracy and macro-F1 for guidance detection, topic classification, and end-to-end evaluation. Macro-F1 gives equal weight to each class, which is important because several topics are rare; the supplementary material provides additional guidance metrics and confidence intervals.

The primary goal of the manual annotation was to verify whether each conversation seeks legal guidance. Because guidance requests span many legal topics, the gold evaluation set provides more limited evidence for topic evaluation: its 84 guidance-positive conversations cover 12 of the 14 topics, with no examples of ``Torts \& Civil Liability'' or ``\textsc{Other},'' and five topics have five or fewer examples. We compute gold topic macro-F1 over only the 12 observed classes and provide bootstrap intervals and paired significance tests in the supplementary material.

\section{How Users Seek Legal Guidance}

Of the 2,122 conversations in the silver-labeled dataset, 1,367 (64.4\%) contain some form of legal content. Among those 1,367 conversations, 972 (71.1\%) seek personalized or practical legal guidance; these guidance requests account for 45.8\% of the dataset. The remaining 395 conversations discuss legal issues without seeking guidance, and 755 contain no legal content. These percentages should not be interpreted as prevalence estimates for WildChat because the sampling process deliberately included more conversations initially identified as legal.

\noindent\textbf{Request types.} Among the 1,367 legal-content conversations, ``rights-and-options advice'' is the most common primary request type (712; 52.1\%), followed by ``information or explanation'' (293; 21.4\%), ``procedural how-to'' (180; 13.2\%), ``document drafting'' (137; 10.0\%), ``other request'' (26; 1.9\%), and ``document review'' (19; 1.4\%). ``Rights-and-options advice'' concerns the user's own situation (e.g., ``Can my landlord evict me for this?''); ``information or explanation'' asks how a legal concept or system works (e.g., ``What is the difference between civil and criminal law?''); and ``procedural how-to'' requests ask for concrete steps, forms, or deadlines (e.g., ``How do I file an appeal?''). ``Document drafting'' asks the model to produce or revise legally operative text (e.g., ``Draft a termination clause''), whereas ``document review'' asks it to analyze an existing document (e.g., ``What does this clause mean?''). The residual ``other request'' category covers legal-content requests outside these groups (e.g., ``Create a quiz about contract law''). Requests to create quizzes or similar study materials fall under this residual category, while requests to explain legal concepts fall under information or explanation.

A conversation can receive more than one request-type label. When both primary and secondary request-type labels are counted, 62.5\% of conversations include ``rights-and-options advice'' and 36.6\% include ``information or explanation.'' Thus, users most often ask what they can or should do in their particular situation. Explanations also appear in more than one third of conversations, often alongside a request for practical guidance.

\noindent\textbf{Topics and guidance onset.} Among the 972 guidance-seeking conversations, ``Immigration \& Citizenship'' is the largest primary topic (14.1\%), followed by ``Criminal Law'' (11.4\%), ``Intellectual Property'' (11.3\%), ``Business \& Commercial'' (11.0\%), and ``Tax Law'' (10.8\%). The distribution has a long tail: ``Civil Rights \& Constitutional'' accounts for 0.3\% of primary labels and ``Torts \& Civil Liability'' for 1.2\%. In total, 25.3\% of guidance-positive conversations receive more than one topic. Secondary labels substantially increase the apparent coverage of cross-cutting domains such as ``Government \& Administrative'' and ``Business \& Commercial'' (Table~\ref{tab:topic-distribution}).

\begin{table}[t]
\centering
\footnotesize
\setlength{\tabcolsep}{6pt}
\begin{tabular}{lrrr}
\toprule
Topic & Primary $n$ & Primary \% & Any \% \\
\midrule
Immigration \& Citizenship & 137 & 14.1 & 14.6 \\
Criminal Law & 111 & 11.4 & 11.9 \\
Intellectual Property & 110 & 11.3 & 12.0 \\
Business \& Commercial & 107 & 11.0 & 17.9 \\
Tax Law & 105 & 10.8 & 13.3 \\
Employment \& Labor & 87 & 9.0 & 10.0 \\
Housing \& Property & 81 & 8.3 & 9.5 \\
Government \& Administrative & 73 & 7.5 & 15.2 \\
Consumer \& Personal Finance & 54 & 5.6 & 8.2 \\
Family \& Estates & 51 & 5.2 & 6.6 \\
Data Privacy \& Technology & 36 & 3.7 & 5.8 \\
Torts \& Civil Liability & 12 & 1.2 & 2.4 \\
\textsc{Other} & 5 & 0.5 & 0.6 \\
Civil Rights \& Constitutional & 3 & 0.3 & 2.2 \\
\bottomrule
\end{tabular}
\caption{Legal-topic distribution among the 972 silver-labeled guidance-positive conversations. ``Primary'' counts the predominant topic; ``Any'' also counts secondary-topic assignments and therefore does not sum to 100\%.}
\label{tab:topic-distribution}
\end{table}

Conversations contain a median of 2 user turns, and user turns contain a median of 51 words. Guidance first appears after the opening user turn in 305 of 972 positives (31.4\%), so 68.6\% begin with guidance; these onset fields are not independently validated and are used only descriptively. The silver annotator flags medium or high uncertainty for 20.7\% of conversations it labels guidance-positive, compared
with 7.6\% of the remaining conversations. We did not test
whether these flags predict human corrections or classifier
errors for this study.

\section{Classification Models and Early Prediction}

\subsection{Experimental setup}

We compare two classifier families: logistic regression using
word- and character-level TF--IDF features, and ModernBERT
encoders in base and large configurations
\cite{pedregosa2011scikit,warner2024modernbert}.
We chose these models with consumer-device use in mind and
aimed for an approach that could process entire datasets quickly. TF--IDF with
logistic regression provides a lexical baseline, while
ModernBERT provides a contextual language-model approach.
The base and large configurations allow us to examine whether
increased model size consistently improves classification
performance. All classifiers receive only user turns in
chronological order.

The training and validation splits have similar guidance rates (46.0\% and 46.2\%), and both contain examples from all 14 topics. Because the topic distribution is highly imbalanced, we use class-weighted topic training and macro-averaged evaluation.

To support guidance detection as a conversation unfolds, we also train binary prefix models. A conversation prefix contains all user messages up to a given turn, whereas the full-conversation models are trained only on complete conversations. For conversations in which guidance seeking begins at turn $k$, prefixes before turn $k$ are labeled negative, and those at or after turn $k$ are labeled positive; conversations without guidance seeking contribute only negative examples. All prefixes from a conversation remain within the same split, and each is weighted by the inverse of the conversation's turn count, so every conversation contributes a total weight of 1. This setup applies only to guidance detection; we do not train a prefix topic classifier.

\noindent\textbf{Training configurations.} Table~\ref{tab:training-config} summarizes the features, optimization settings, and validation rules for the full-conversation and prefix models. The frozen prefix splits contain 4,865 training prefixes from 1,632 conversations and 912 validation prefixes from 290 conversations, including 2,042 and 376 positive prefixes, respectively.

\begin{table}[t]
\centering
\scriptsize
\setlength{\tabcolsep}{3pt}
\begin{tabular}{@{}p{0.15\textwidth}p{0.25\textwidth}p{0.30\textwidth}p{0.22\textwidth}@{}}
\toprule
System & Input and representation & Training configuration & Validation selection \\
\midrule
Full TF--IDF
& Complete user turns; word 1--2-grams (150k) and character 3--5-grams (200k)
& Class-balanced logistic regression; selected $C=1$ from $\{0.25,1,4\}$
& Guidance threshold 0.51 selected on validation data \\
\addlinespace
Full ModernBERT
& Complete user turns; base and large encoders; maximum length 4,096
& Batch size 32, weight decay 0.01, seed 42; guidance: 3 epochs at $5\times10^{-5}$; class-weighted topic: 10 epochs at $6\times10^{-5}$
& Checkpoint selected by macro-F1; guidance thresholds: 0.26 (base), 0.38 (large) \\
\addlinespace
Prefix TF--IDF
& User-turn prefixes; word 1--2-grams and character-within-word 3--5-grams
& Logistic regression with $C=0.25$; conversation-level weighting
& Weighted macro-F1; threshold 0.42 \\
\addlinespace
Prefix ModernBERT
& User-turn prefixes; base and large encoders; maximum length 4,096
& Batch size 32, weight decay 0.01, seed 42; 3 epochs at $5\times10^{-5}$; conversation-level weighting
& Weighted validation loss; thresholds: 0.51 (base), 0.36 (large) \\
\bottomrule
\end{tabular}
\caption{Training configurations. Thresholds apply to guidance detection. Prefix thresholds maximize conversation-weighted macro-F1, using positive-class F1 as a tie-breaker.}
\label{tab:training-config}
\end{table}

\noindent\textbf{Resources.} The \href{https://huggingface.co/collections/AmirMohseni/legal-guidance-in-the-wild}{Legal Guidance in the Wild Hugging Face collection} is the consolidated entry point for the datasets, model checkpoints, training details, human-reviewed gold evaluation set, and evaluation artifacts.

\subsection{Evaluation}

We calibrated the guidance models on the silver validation split using the thresholds in Table~\ref{tab:training-config} and froze all model-selection decisions before evaluating on the gold evaluation set. The final-turn results use this locked evaluation; the exploratory turn-1 and turn-2 analysis applies the same frozen prefix checkpoints and thresholds to shorter inputs. We report 95\% bootstrap intervals from 2,000 conversation-level resamples and use exact paired McNemar tests to compare system errors. Full reliability procedures and results are provided in the supplementary material.

\section{Results}
\label{sec:results}

Final-turn results come from the locked evaluation on the 200-conversation gold evaluation set. The initial-turn results come from a subsequent exploratory analysis with the frozen prefix models. Full per-class topic scores are provided in the supplementary material.

\subsection{Stage 1: Guidance detection}

\begin{table}[t]
\centering
\footnotesize
\setlength{\tabcolsep}{8pt}
\begin{tabular}{lrr}
\toprule
Model & Acc. & Macro-F1 \\
\midrule
\multicolumn{3}{l}{\textit{Full conversation}} \\
\quad TF--IDF + logistic regression & .840 & .839 \\
\quad ModernBERT-base & .815 & .813 \\
\quad ModernBERT-large & \textbf{.845} & \textbf{.845} \\
\midrule
\multicolumn{3}{l}{\textit{Prefix models (final turn)}} \\
\quad TF--IDF + logistic regression & .825 & .823 \\
\quad ModernBERT-base & .830 & .828 \\
\quad ModernBERT-large & \textbf{.860} & \textbf{.859} \\
\bottomrule
\end{tabular}
\caption{Guidance detection on the 200-conversation gold evaluation set. Prefix models are scored after the final user turn. Full metrics and bootstrap intervals are provided in the supplementary material. Bold marks the best value within each block.}
\label{tab:gold-guidance}
\end{table}

ModernBERT-large has the highest full-conversation guidance scores, with accuracy and macro-F1 both at .845 (Table~\ref{tab:gold-guidance}). The margin over the lexical baseline is small: TF--IDF reaches .840 accuracy, a difference of one conversation in 200. ModernBERT-base has the lowest accuracy and macro-F1.

ModernBERT-large has the highest recall, identifying 79 of 84 guidance-positive conversations (.940), compared with 75 for TF--IDF and 71 for ModernBERT-base, but it also produces 26 false positives.

The prefix ModernBERT-large model has the highest final-turn guidance scores, with .860 accuracy and .859 macro-F1. Table~\ref{tab:gold-prefix-turns} provides an exploratory view of initial prefixes. From the opening turn, TF--IDF has the highest macro-F1 (.826); among conversations with at least two turns, ModernBERT-large has the highest turn-2 macro-F1 (.865). Because the two columns cover different conversation sets and use eventual conversation-level guidance labels, they do not measure improvement over time or correctness of guidance onset.

\begin{table}[t]
\centering
\footnotesize
\setlength{\tabcolsep}{6pt}
\begin{tabular}{lrrrr}
\toprule
& \multicolumn{2}{c}{Turn 1 ($n=200$)} & \multicolumn{2}{c}{Turn 2 ($n=104$)} \\
\cmidrule(lr){2-3}\cmidrule(lr){4-5}
Model & Acc. & Macro-F1 & Acc. & Macro-F1 \\
\midrule
TF--IDF + logistic regression & \textbf{.830} & \textbf{.826} & .798 & .798 \\
ModernBERT-base & .800 & .792 & .808 & .808 \\
ModernBERT-large & .825 & .819 & \textbf{.865} & \textbf{.865} \\
\bottomrule
\end{tabular}
\caption{Exploratory initial-turn prediction results. Turn 1 uses all conversations in the gold evaluation set; Turn 2 uses the 104 conversations containing at least two user turns. Labels indicate whether the complete conversation eventually seeks guidance, not whether guidance has begun by that turn. The columns use different evaluation sets and should not be interpreted as within-model improvement.}
\label{tab:gold-prefix-turns}
\end{table}

\subsection{Topic classification}

\begin{table}[t]
\centering
\footnotesize
\setlength{\tabcolsep}{3.5pt}
\begin{tabular}{lrrlrrrrrr}
\toprule
& \multicolumn{3}{c}{Topic only$^{\dagger}$} & \multicolumn{2}{c}{End-to-end} & \multicolumn{4}{c}{Cascade errors} \\
\cmidrule(lr){2-4}\cmidrule(lr){5-6}\cmidrule(lr){7-10}
Model & Acc. & Macro-F1 & \multicolumn{1}{c}{95\% CI} & Acc. & Macro-F1 & FP & FN & Topic & All \\
\midrule
TF--IDF + logistic regression & .679 & .559 & [.441, .632] & .740 & .527 & 23 & 9 & 20 & 52 \\
ModernBERT-base & .750 & .646 & [.537, .715] & .725 & .544 & 24 & 13 & 18 & 55 \\
ModernBERT-large & \textbf{.762} & \textbf{.671} & [.538, .750] & \textbf{.755} & \textbf{.593} & 26 & 5 & 18 & \textbf{49} \\
\bottomrule
\end{tabular}
\caption{Topic-only classification and end-to-end routing on the gold evaluation set. $^{\dagger}$Topic-only evaluation uses gold guidance labels for the 84 guidance-positive conversations, which represent only 12 of the 14 classes; that fixed set is used for macro-F1 and its bootstrap intervals. End-to-end scores use all 200 conversations and the 13 classification outcomes observed in the evaluation set. Cascade errors comprise false-positive (FP) routing, false-negative (FN) routing, and wrong-topic predictions after correct positive routing (Topic).}
\label{tab:gold-topic-cascade}
\end{table}

Both ModernBERT models score higher than TF--IDF on topic classification (Table~\ref{tab:gold-topic-cascade}). Using gold guidance labels, ModernBERT-large correctly classifies 64 of 84 conversations, ModernBERT-base 63, and TF--IDF 57. Their macro-F1 scores are .671, .646, and .559, respectively. In end-to-end evaluation, ModernBERT-large routes 151 of 200 conversations correctly, compared with 148 for TF--IDF and 145 for ModernBERT-base.

The error counts explain why the topic-only and end-to-end rankings differ. ModernBERT-base performs well on topic classification but misses 13 guidance-positive conversations at Stage 1. ModernBERT-large misses only five, giving it the best end-to-end result despite producing 26 false positives. Across the three cascades, Stage-1 classification accounts for 31--37 of the 49--55 total errors, and false positives are the largest single error category (23--26 cases). Wrong-topic predictions remain substantial (18--20 cases), so improvements at either stage could change end-to-end performance.

Per-class performance varies widely, particularly for topics with few gold examples. Five of the 12 observed topics have five examples or fewer, so these results are descriptive rather than stable class-level estimates.

\subsection{Statistical reliability and silver-label effects}
\label{sec:reliability}

No exact paired comparison reaches $p < .05$, and the bootstrap macro-F1 intervals overlap. The clearest performance advantage of dense models over TF--IDF is in topic classification, where ModernBERT-base and ModernBERT-large improve macro-F1 by 8.7 and 11.2 points, respectively.

Model size was not predictive of performance across the tasks. The base model led the large model by 2.9 macro-F1 points on the silver validation set for topic classification, while the large model led by 2.5 points on the gold topic task. The large model also scored highest on full-conversation guidance detection and end-to-end classification. These mixed results do not show a consistent advantage for either model size.

\section{Discussion}

The results suggest that ModernBERT helps most with topic classification. Guidance requests often contain clear words and phrases associated with options, forms, deadlines, or document drafting, which may explain why TF--IDF remains competitive. Topic classification requires separating related and sometimes overlapping areas of law, and both ModernBERT models perform better on this task. However, the base and large models perform differently across the silver and gold datasets, and the differences between them are not statistically significant.

The errors also show a practical trade-off. ModernBERT-large misses the fewest guidance requests but also incorrectly classifies the most non-guidance conversations as guidance requests. This may be acceptable for a simple caution notice, but it would increase workload if flagged conversations are sent to human reviewers or specialists. The decision threshold should therefore match the intended use. Moreover, 25.3\% of guidance-seeking conversations span multiple topics, while the classifier assigns just one. Consequently, some topic classification errors may stem from the presence of multiple topics, rather than incorrect predictions. 

Several design choices limit the scope of our findings. Training and development labels come from one language model, and reviewers assessed pre-annotated labels, which may have introduced anchoring effects \cite{berzak2016anchoring}. Both reviewers had computer-science backgrounds, and the topic labels were not independently checked by a legal expert. Request types, secondary topics, and uncertainty flags were not human-validated. The enriched sampling design supports classifier development but not prevalence estimates for WildChat. The gold evaluation set contains only 84 guidance-positive conversations across 12 observed topics, and the models were trained using a single random seed. The taxonomy's focus on English, its jurisdiction-agnostic design, and the reduction of multi-topic requests to one primary routing label further limit transfer to other settings. Guidance onset was not human-validated, so the initial-turn analysis measures whether eventual guidance seeking is predictable from initial turns rather than whether a model detects its true onset.

The conversations may describe sensitive legal circumstances despite WildChat's anonymization procedures \cite{zhao2024wildchat}. In this paper, we report aggregate results and do not reproduce conversation excerpts. The classifier identifies requests that may need safeguards or specialized treatment; it does not determine whether a claim is valid or an answer is correct. Calibration, user testing, and access to human support would therefore be necessary before deployment.

\section{Conclusion}

We introduce a taxonomy and classification benchmark for studying
legal guidance seeking in general-purpose LLM conversations.
In this enriched sample, silver labels identify guidance
seeking in 45.8\% of conversations. Rights-and-options advice
is the most common primary request type among legal-content
conversations, and 25.3\% of guidance-positive conversations
span multiple topic labels.

On the gold evaluation set, ModernBERT-large achieves the best performance overall; however, the gains are not uniform: TF--IDF performs similarly to dense models on binary guidance detection, whereas dense models lead on topic classification. No paired comparison reaches significance at this sample size, so the score differences should be interpreted cautiously.

Future work should extend this taxonomy, test its robustness across languages and legal systems, and explore how classifier outputs might inform real-time caution notices or routing decisions in deployed conversational AI systems. More broadly, we hope the public release of the annotated dataset, taxonomy, and evaluation code supports both further research into legal help-seeking behavior and the development of practical safeguards in deployed systems.

\section*{Declaration of generative AI use}

OpenAI's GPT-5.6 Terra was used during taxonomy development,
and GPT-5.6 Sol was used to generate the final silver
annotations, as described in Section~\ref{sec:data}.
GPT-5.6 Sol was also used to refine grammar and correct
LaTeX sections of the manuscript. The authors reviewed
and edited the resulting content and take responsibility
for the final publication.

\bibliographystyle{vancouver}
\bibliography{references}

\end{document}